\documentclass[letterpaper, 10 pt, conference]{ieeeconf}  

\IEEEoverridecommandlockouts                              

\usepackage{amsmath} 
\usepackage{amssymb}  
\usepackage{url}
\usepackage{graphicx}
\usepackage{cite}
\usepackage{algorithm}
\usepackage{algorithmic}
\usepackage{multirow}
\usepackage{lipsum}

\title{\LARGE \bf
Robot-Assisted Navigation for Visually Impaired through\\Adaptive Impedance and Path Planning
}

\author{Pietro Balatti$^{1,*}$, Idil Ozdamar$^{1,2,*}$, Doganay Sirintuna$^{1,2,*}$, Luca Fortini$^1$,\\Mattia Leonori$^1$, Juan M. Gandarias$^{3}$, and Arash Ajoudani$^1$
\vspace{-4mm}
\thanks{$^*$ Contributed equally to this work.}
\thanks{$^1$HRI$^2$ Lab, Istituto Italiano di Tecnologia, Genoa, Italy. {\tt\small pietro.balatti@iit.it}, {$^2$ Dept. of Informatics, Bioengineering, Robotics, and System Engineering. University of Genoa, Genoa, Italy.}, {$^3$ Robotics and Mechatronics lab, Systems Engineering and Automation Department, University of Malaga, Malaga, Spain.}}
\thanks{This work was supported  in part by the European Union’s Horizon 2020 research and innovation programme under Grant Agreement No. 871237 (SOPHIA) and by the project "RAISE - Robotics and AI for Socio-economic Empowerment” (supported by European Union - NextGenerationEU).}%
}

\begin{document}

\maketitle
\thispagestyle{empty}
\pagestyle{empty}

\begin{abstract}
This paper presents a framework to navigate visually impaired people through unfamiliar environments by means of a mobile manipulator. The Human-Robot system consists of three key components: a mobile base, a robotic arm, and the human subject who gets guided by the robotic arm via physically coupling their hand with the cobot's end-effector. These components, receiving a goal from the user, traverse a collision-free set of waypoints in a coordinated manner, while avoiding static and dynamic obstacles through an \textit{obstacle avoidance} unit and a novel \textit{human guidance} planner. With this aim, we also present a \textit{legs tracking} algorithm that utilizes 2D LiDAR sensors integrated into the mobile base to monitor the human pose. Additionally, we introduce an \textit{adaptive pulling} planner responsible for guiding the individual back to the intended path if they veer off course. This is achieved by establishing a target arm end-effector position and dynamically adjusting the impedance parameters in real-time through a \textit{impedance tuning} unit. To validate the framework we present a set of experiments both in laboratory settings with 12 healthy blindfolded subjects and a proof-of-concept demonstration in a real-world scenario.
\end{abstract}

\section{Introduction}\label{sec:introduction}

In recent decades, collaborative robotics has experienced noteworthy advancements. We have transitioned from confining industrial robots within safety enclosures, where any physical interaction with humans was banned, to a scenario where robots seamlessly collaborate with humans, even when contacts occur~\cite{hagele2016industrial, ajoudani2018progress}.  
In many contemporary applications, human-robot physical contact is not only allowed but required in various tasks. Rehabilitation applications are a representative illustration of this premise~\cite{qassim2020review, zhang2020development, ruiz2021upper}. Advanced flexible manufacturing also demands physical Human-Robot Interaction (pHRI) for many Human-Robot Collaboration (HRC) tasks, such as collaborative transportation of objects~\cite{sirintuna2023object, sirintuna2023carrying} or conjoined actions for heavy tasks~\cite{giammarino2022super}. Search-and-rescue robotics also benefits from it for casualties extraction in disaster areas~\cite{choi2019development} or limbs manipulation~\cite{gandarias2019underactuated}. Nevertheless, assistive and healthcare robotics is presumably the most interesting application, where the number of works employing this approach is growing notably~\cite{ding2014giving, nakamura2018soft, liu2018biomimetic, fitter2020exercising, su2021physical, ruiz2022improving}.

\begin{figure}[!t]
\centering
\includegraphics[width=0.84\columnwidth]{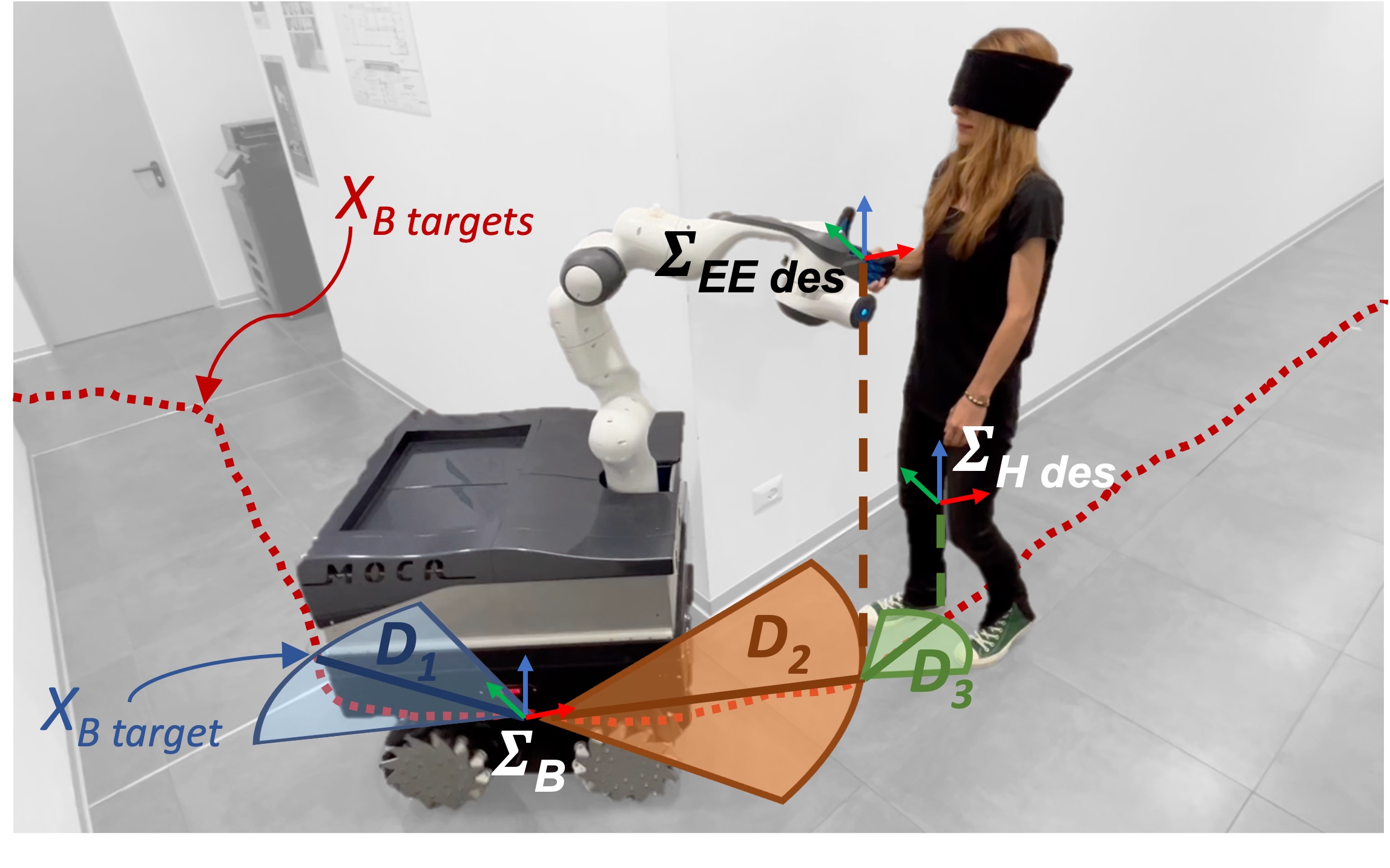}
\vspace{-4mm}
\caption{Visually impaired people are navigated through an unfamiliar environment by means of a mobile manipulator. By tracking the human pose in the workspace and tuning online the impedance parameters at the point of interaction, the framework ensures a collision-free path planning.}
\label{fig:intro}
\vspace{-6mm}
\end{figure}

The development of assistive robotic systems to guide the visually impaired is an important application that could benefit significantly from advances in this field~\cite{hong2022development}. Some works have presented robotic solutions to this particular topic. In~\cite{wakita2011human, ye2016co, van2019robotic, slade2021multimodal}, different versions of robotized white canes that can assist in visually impaired guidance are proposed. More recently, the use of legged or dog-like robots is being studied~\cite{xiao2021robotic, due2023guide, chen2023quadruped}. This application has also been recently considered with aerial robots~\cite{tognon2021physical}. However, to the best of the authors' knowledge, the use of a mobile robot with a manipulator for this application has yet to be explored. Exploiting the loco-manipulation capabilities of a mobile manipulator is crucial for guidance in labyrinthine environments or corridors with recesses or tight turns. 

\begin{figure*}[!t]
\centering
\includegraphics[width=0.8\textwidth]{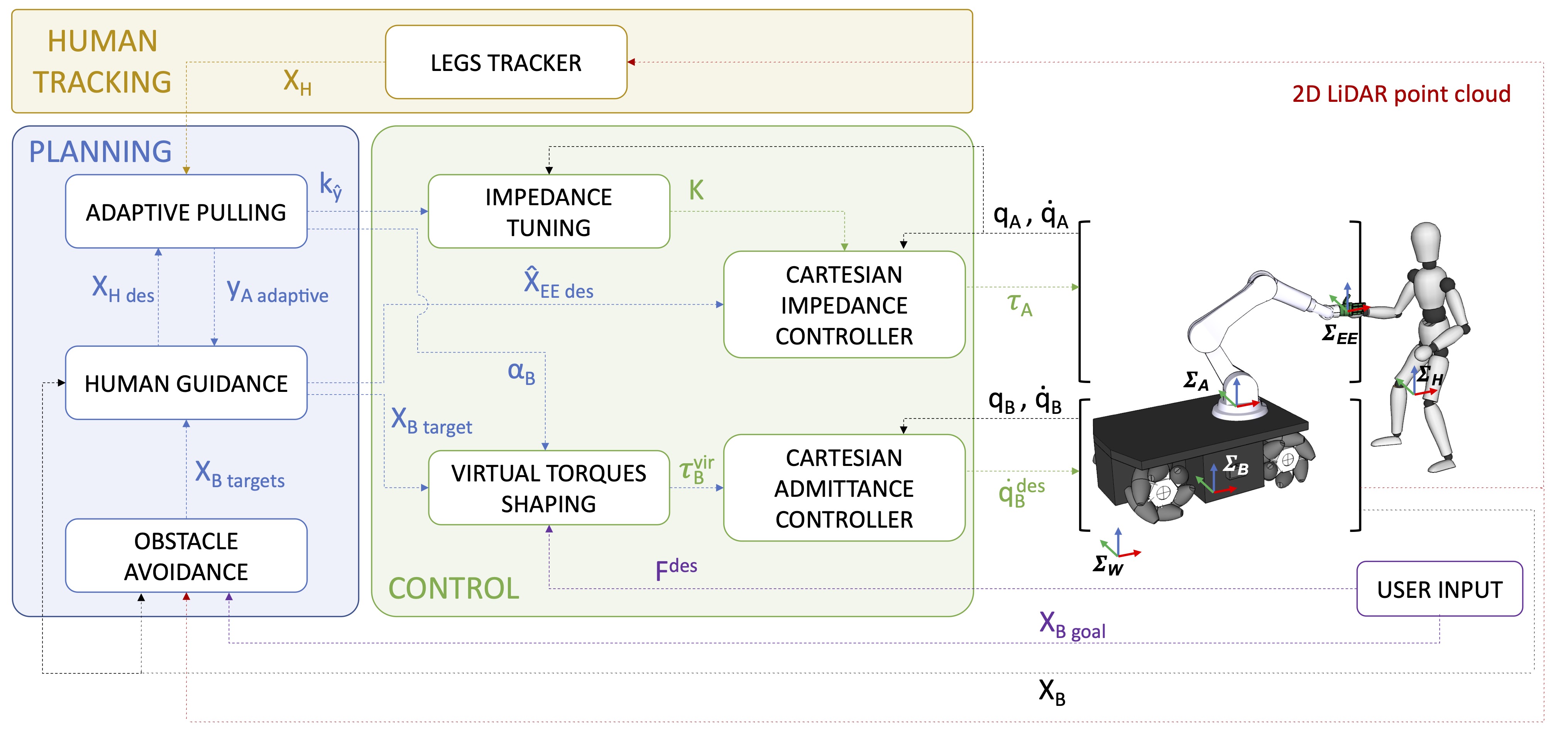}
\vspace{-3mm}
\caption{The software architecture of the presented framework includes human tracking (yellow), planning algorithms (blue), and control methods (green). Every unit functions as a ROS node, and information is communicated through ROS messages along the illustrated ROS topics (indicated by dotted lines).}
\label{fig:software-architecture}
\vspace{-6mm}
\end{figure*}

In this work, we propose a novel pHRI framework that enables a mobile manipulator to plan and execute a trajectory exploiting the loco-manipulation skills to navigate a visually impaired person in an unknown environment (see Fig.~\ref{fig:intro}). One significant advantage of personal mobile robots is that they can provide greater assistance in daily living activities, due to their manipulation capacity.
For instance, the mobile base can be used for carrying heavy loads, and the robotic arm can be exploited to open doors or pick items that lie out of the human workspace. This framework aims to enhance the independence of visually impaired people, and this platform is meant to be a constant helper that can accompany the users to carry out their daily activities.

The core of this work and its main novelty relies on a guidance planner that ensures a collision-free path for the HRI system composed by the robot and the physically coupled user, exploiting an adaptive pulling vector that through physical guidance indicates to the user the path to be followed. This is achieved by tracking the human pose in the robot surroundings, monitoring the leg movements, and commanding the arm end-effector pose and impedance parameters. 

The default behavior of the robotic arm is to keep a compliant profile on the lateral axis until the human does not veer off the desired path. In fact, robots with lower stiffness are often perceived as safer by humans~\cite{rubagotti2022perceived}. Humans tend to feel more secure when they believe the robot can adapt to unexpected situations without causing harm. This can be relevant to our study since undesired end-effector movements may be generated due to the external disturbances, planning uncertainties, and imperfect de-coupling of the base and arm movements from the end-effector trajectories.  Hence, when a robot is compliant and can yield to external forces, it reduces the risk of accidental collision or unexpected interaction. Lower stiffness has also shown to increase the feeling of dominance by the individuals interacting with robots~\cite{fitter2020does}. Conversely, robots with high stiffness can be seen as potentially more dangerous. If a robot is rigid and unyielding, it may not respond well to external forces, increasing the risk of injury or discomfort for humans, simultaneously reducing the feeling of dominance. As a result, individuals might be cautious or hesitant when interacting with such robots.

\vspace{-2mm}
\section{Method}\label{sec:method}
The main purpose of the proposed framework is to navigate visually impaired people through an unknown environment, thanks to the assistance provided by a mobile manipulator. When the human subjects couple their hand with the anthropomorphic robotic hand mounted as the end-effector, the cobot starts pulling the person towards a predefined goal while avoiding static and dynamic obstacles. To do so, an adaptive pulling strategy has been designed to shape both the desired end-effector pose and the impedance parameter of the Cartesian impedance controller embedded in the arm. This leads to a readjustment of the human pose in the case an individual deviates from a safe, i.e., collision-free, track shaped by the framework.

The required theoretical and technological components to build such a framework are integrated into 8 modules, that can be subdivided into three main categories: human tracking, planning algorithms, and control methods, as illustrated in Fig.~\ref{fig:software-architecture}. 
The employed control techniques include (1) a \textit{Cartesian impedance controller} whose impedance parameters can be tuned online by the (2) \textit{impedance tuning} unit, (3) a \textit{Cartesian admittance controller} that commands the desired velocity to the robot base, receiving as input the virtual torques generated by the (4) \textit{virtual torques shaping} unit. The human tracking consists of (5) a \textit{legs tracker} unit that, by using 2D point cloud data, detects the human location with respect to the robot. The planning algorithms involve (6) an \textit{obstacle avoidance} planner that dynamically generates a trajectory for the mobile base able to ensure a collision-free path, (7) an \textit{adaptive pulling} unit that, based on the desirable human pose and their actual one, shapes a 2D pulling vector that determines if (8) the \textit{human guidance} planner unit needs to steer the robotic manipulator to bring back the human subject on the desired generated track.

\vspace{-2mm}
\subsection{Cartesian impedance controller}\label{subsec:cartesian-impedance-controller}
In HRC tasks, Cartesian impedance control techniques have demonstrated the ability to guarantee safe interactions between the cobot and the human counterpart~\cite{ajoudani2018progress}, being able to achieve any arbitrary dynamic behavior at the robot end-effector \cite{schindlbeck2015unified,ajoudani2017choosing}. This control technique relies on torque sensing and actuation, with the vector of robotic arm joint torques $\boldsymbol{\tau}_A \in \mathbb{R}^{n}$ calculated as follows:
\begin{equation}\vspace{-2mm}
  \boldsymbol{\tau}_A =  \boldsymbol{M(q)\ddot{q}} + \boldsymbol{C(q,\dot{q})\dot{q}} + \boldsymbol{g(q)}   + \boldsymbol{\tau}_{ext},
\end{equation}
\begin{equation}\vspace{-2mm}
  \boldsymbol{\tau}_{ext} =  \boldsymbol{J(q)}^T \boldsymbol{F}_c + \boldsymbol{\tau}_{st},
\end{equation}
where $n$ is the number of joints, $\boldsymbol{q} \in \mathbb{R}^{n}$ is the joint angles vector, $\boldsymbol{J} \in \mathbb{R}^{6 \times n}$ is the robot arm Jacobian matrix, $\boldsymbol{M} \in \mathbb{R}^{n \times n}$ is the mass matrix, $\boldsymbol{C} \in \mathbb{R}^{n \times n}$ is the Coriolis and centrifugal matrix, $\boldsymbol{g} \in \mathbb{R}^{n}$ is the gravity vector and $\boldsymbol{\tau}_{ext}$ is the external torque vector. $\boldsymbol{F}_c$ represents the forces vector in the Cartesian space and $\boldsymbol{\tau}_{st}$ the second task torques projected onto the null-space of $\boldsymbol{J}$. 

Forces $\boldsymbol{F}_c \in \mathbb{R}^{6}$ are calculated as follows:
\vspace{-1mm}
\begin{equation}\vspace{-1mm}
  \boldsymbol{F}_c =  \boldsymbol{K}_c (\boldsymbol{X}_d - \boldsymbol{X}) + \boldsymbol{D}_c (\boldsymbol{\dot{X}}_d - \boldsymbol{\dot{X}}),
\end{equation}

\noindent where $\boldsymbol{K}_c \in \mathbb{R}^{6 \times 6}$ and $\boldsymbol{D}_c \in \mathbb{R}^{6 \times 6}$ represent respectively the Cartesian stiffness and damping matrix, $\boldsymbol{X}_d$ and $\boldsymbol{X} \in \mathbb{R}^{6}$ the Cartesian desired and actual poses, $\boldsymbol{\dot{X}}_d$ and $\boldsymbol{\dot{X}} \in \mathbb{R}^{6}$ their corresponding velocity profiles.

\subsection{Impedance tuning}\label{subsec:impedance-tuning}\vspace{-2mm}
In~\cite{balatti2020method}, we introduced the principles of a self-tuning impedance controller that allows the robot to precisely track the desired trajectory along the motion vector, and at the same time grants the flexibility to respond in a compliant way along the other directions in order to gently adapt to external unintended disturbances (e.g., obstacles). Building upon the basis of this work, hereafter we introduce an extended version of the algorithm that is capable of independently regulating the impedance parameters along the three axes of the impedance ellipsoid located at the robot end-effector frame, whose axes are oriented in the direction of the motion ($\boldsymbol{a_x}$), perpendicular to the direction of the motion and laying on a plane parallel to the ground ($\boldsymbol{a_y}$), and orthogonal to the ground ($\boldsymbol{a_z}$). In the presented framework, the independent impedance regulation along the aforementioned axes allows the robot (i) to keep a stiff profile in the principal direction of the movement guaranteeing the pulling of the human subject towards the final goal through $\boldsymbol{a_x}$, (ii) to leave the freedom of motion along the lateral axis when the human pose is tracked to be within the collision-free path or to make it stiffer when the robot needs to redirect the human on the right track (see Sec.~\ref{subsec:adaptive-pulling}) through $\boldsymbol{a_y}$, and (iii) to set a neutral profile on the vertical axis to leave to the user the possibility to slightly adapt the position of the hand in the vertical direction through $\boldsymbol{a_z}$.

To this end, the translational component of the Cartesian stiffness and damping matrices (symmetric and positive definite) $\boldsymbol{K}_{c,t} \in \mathbb{R}^{3 \times 3}$, $\boldsymbol{D}_{c,t} \in \mathbb{R}^{3 \times 3}$ are defined as follows:
\begin{equation}
\begin{aligned}\label{eq:selftuning}\vspace{-4mm}
    \boldsymbol{K}_{c,t} = \boldsymbol{U}\boldsymbol{\Sigma}_{k}\boldsymbol{U}^{T}, \qquad\qquad& \boldsymbol{D}_{c,t} = \boldsymbol{U}\boldsymbol{\Sigma}_{d}\boldsymbol{U}^{T},
\end{aligned}
\end{equation}
where the diagonal matrix $\boldsymbol{\Sigma}_{k} \in \mathbb{R}^{3 \times 3}$ and $\boldsymbol{\Sigma}_{d} \in \mathbb{R}^{3 \times 3}$ are the desired stiffness and damping factors along the three axis of the impedance ellipsoid. $\boldsymbol{U} \in \mathbb{R}^{3 \times 3}$ is the orthonormal basis whose columns represent the axes defined above as $\boldsymbol{a_x} \in \mathbb{R}^{3}$, $\boldsymbol{a_y} \in \mathbb{R}^{3}$, and $\boldsymbol{a_z} \in \mathbb{R}^{3}$, and calculated as:
\begin{equation}\label{eq:u-selftuning}\vspace{-2mm}
\boldsymbol{U} = [\begin{matrix}
\boldsymbol{\hat{a}_x}, \ \boldsymbol{\hat{a}_y}, \ \boldsymbol{\hat{a}_z}
\end{matrix}]
\end{equation}
\begin{equation}\label{eq:ax}\vspace{-2mm}
\boldsymbol{a_x} = \boldsymbol{x_{d,t}} - \boldsymbol{x_{d,t-1}},
\end{equation}
\begin{equation}\label{eq:ay}\vspace{-2mm}
\boldsymbol{a_y} = [-\boldsymbol{a_x}(y) \ \boldsymbol{a_x}(x) \ 0]^T,
\end{equation}
\begin{equation}\label{eq:az}\vspace{-3mm}
\boldsymbol{a_z} = \boldsymbol{a_x} \times \boldsymbol{a_y},
\end{equation}
where $\hat{}$ denotes that the vectors have been normalized, $\boldsymbol{x_{d,t}}$ and $\boldsymbol{x_{d,t-1}}$ are the translational component of $\boldsymbol{X}_d$ at the current and former control loop, and $\times$ represents the cross product notation. $\boldsymbol{\Sigma}_{k}$ and $\boldsymbol{\Sigma}_{d}$ are defined by:
\begin{equation}
\begin{aligned}\label{eq:sigma}\vspace{-4mm}
    \boldsymbol{\Sigma}_{k} = \textrm{diag}({k_x},{k_y},{k_z}), \qquad& \boldsymbol{\Sigma}_{d} = \textrm{diag}({d_x},{d_y},{d_z}),
\end{aligned}
\end{equation}
where ${k_x}$, ${k_y}$, and ${k_z}$ represent the desired stiffness values to be projected on the $\boldsymbol{U}$ basis axes. The corresponding damping values are computed as $d_i = 2\zeta\sqrt{k_i}$ with $i \in (x,y,z)$ \cite{albu2003cartesian}.

\subsection{Cartesian admittance controller}\label{subsec:cartesian-admittance-controller}
Robotic mobile platforms can be controlled by means of a velocity-based control and a high gain can be set in the low-level velocity controller. This means that the dynamics of the mobile platform can be omitted and any external dynamic effect from the manipulator can be ignored. With the aim of commanding the base movements based on the user interaction with the robot, a force-torque interface is preferred. To map the high level torques computed by a virtual torque controller into suitable velocities for the base, an admittance controller is used. The number of Degrees of Freedom (DoF) $m$ = 3, as the mobile platform can move along the X-axis, the Y-axis, and rotate about the vertical axis (yaw). The dynamics of the mobile platform, with virtual joints $\boldsymbol{q}_{B} \in\mathbb{R}^{m}$ can be described by:
\vspace{-1mm}
\begin{equation}\vspace{-2mm}
\boldsymbol{M}_{adm}\boldsymbol{\ddot{q}}_B^{des} + \boldsymbol{D}_{adm}\boldsymbol{\dot{q}}_B^{des} = \boldsymbol{\tau}_{B}^{vir} + \boldsymbol{\tau}_{B}^{ext},
  \label{eq:mobile_base_dynamics}
\end{equation}
\noindent where $\boldsymbol{M}_{adm} \in\mathbb{R}^{m \times m}$ and $\boldsymbol{D}_{adm} \in\mathbb{R}^{m \times m}$ are the virtual inertial and virtual damping, $\boldsymbol{\dot{q}}_B^{des} \in\mathbb{R}^{m}$ is the desired velocity sent to the mobile platform, $\boldsymbol{\tau}_{B}^{vir} \in\mathbb{R}^{m}$ and $\boldsymbol{\tau}_{B}^{ext} \in\mathbb{R}^{m}$ are the virtual and external torque, respectively. The virtual torques vector is composed by linear forces on the x- and y-axis and rotational torques around the z-axis, and it is defined as:
\vspace{-1mm}
\begin{equation}\vspace{-2mm}
   \boldsymbol{\tau}_{B}^{vir} = [f_x,f_y,\mu_z]^T .
  \label{eq:tau_virt}
\end{equation}

Considering the sampling time $t_s$, the desired velocity $\boldsymbol{\dot{q}}_m^{des}$ can be obtained by substituting $\boldsymbol{\ddot{q}}_m^{des}(t) = t_s^{-1}(\boldsymbol{\dot{q}}_m^{des}(t) - \boldsymbol{\dot{q}}_m^{des}(t-1))$ as:
\vspace{-1mm}
\begin{align}
    \boldsymbol{\dot{q}}_m^{des}(t) =&(t_s^{-1} \boldsymbol{M}_{adm} + \boldsymbol{D}_{adm})^{-1} \\
    &(\boldsymbol{\tau}_{B}^{vir}(t) + \boldsymbol{\tau}_{B}^{ext}(t) + t_s^{-1} \boldsymbol{M}_{adm} \boldsymbol{\dot{q}}_m^{des}(t-1)). \notag
  \label{eq:admittance_controller}
\end{align}

\subsection{Legs tracker}\label{subsec:legs-tracker}
Monitoring the human's position relative to the robotic platform is essential for effective guidance, as depicted in Fig.~\ref{fig:software-architecture}. Mobile robots are usually equipped with a diverse range of sensory systems, offering numerous options. Ultimately, we selected a laser scan-based solution due to its superior performance in terms of frequency, computational efficiency, and angle coverage. The first step consists of filtering the laser point cloud data using a passthrough filter based on user-defined angle and distance ranges to discard noise and background objects. Subsequently, the system employs a DBSCAN (Density-Based Spatial Clustering of Applications with Noise) algorithm to identify clusters within the laser data, representing potential objects or obstacles. It is important to note that this tracking algorithm aims to associate new centroids with previously tracked ones based on their proximity and coherence, allowing the system to follow objects as they move through the sensor's field of view. The coherence threshold parameter, defines the maximum allowed distance between a new centroid and the existing tracked objects. The method  filters out the centroids that do not meet this coherence criterion before registering them as tracked objects. Given that in our experiments, we do not observe significant high accelerations, we decided to set the coherence threshold to a sufficiently restrictive value, such as 0.2 meters. The legs' centroids position are then averaged to define the current human pose frame ($\boldsymbol{\Sigma}_{H}$).

\subsection{Obstacle avoidance}\label{subsec:obstacle-avoidance}
Since the proposed framework is meant to be deployed in unstructured environments with static and dynamic obstacles, it needs to generate a collision-free path for both the mobile base, the robotic arm, and the human subject.

To this end, we decided to implement an obstacle avoidance algorithm capable of preventing collisions with both fixed and moving agents. The method is based on the Robot Operating System (ROS) package ``move\_base'' \cite{movebase}
, using the ROS Global Planner along with the TEB (Timed-Elastic-Band) local planner \cite{rosmann2015planning}. 
This is done by updating a cost map fusing the data perceived by perception sensors, such as lasers and cameras, and the odometry information. Given as input the base goal pose $\boldsymbol{X}_{B\ goal}$, the algorithm generates a vector containing a list of target waypoints, $\boldsymbol{X}_{B\ targets}$, to be reached by the mobile base. It is important to notice that, this package is usually exploited to directly input to the base the velocity commands, while in this framework the \textit{virtual torques shaping} unit is responsible of generating the base virtual torques that are translated into velocities by the \textit{Cartesian admittance controller}.

\subsection{Adaptive pulling}\label{subsec:adaptive-pulling}

The goal of the adaptive pulling module is to prioritize the safety of the human by keeping their position within the robot's collision-free trajectory.
To achieve this, this module generates a vector for pulling the human towards the path to be followed. 
In order to calculate the pulling vector, the desired human pose frame ($\boldsymbol{\Sigma}_{H\ des}$) and the current human pose frame ($\boldsymbol{\Sigma}_{H}$) are utilized.
The translation between these two frames (${\boldsymbol{l}^{H\ des}_{H}}$), with ${l_x}$ and ${l_y}$ corresponding to its x and y components, represents how much the human's actual pose deviates from the desired one. This deviation is exploited to derive the pulling vector as follows:
\vspace{-1mm}
\begin{equation}\vspace{-2mm}
\overrightarrow{\boldsymbol{p}} = [-{l_x}, -{l_y}]^T .
\end{equation}

To ensure safety and prevent the robot from exceeding
its operational limits, the pulling vector is saturated  by the following rule:
\vspace{-2mm}
\begin{equation} \label{eq:v_pull}\vspace{-1mm}
  \overrightarrow{\boldsymbol{p}^{sat}} = \begin{cases}
  \overrightarrow{\boldsymbol{p}}, \ \ \ \ \ \ \ \
 \text{if} \ \ ||\overrightarrow{\boldsymbol{p}}||\leq 1 \\ 
 \dfrac{\overrightarrow{\boldsymbol{p}}}{||\overrightarrow{\boldsymbol{p}}||} ,
  \ \ \ \ \text{else}\end{cases}.
\end{equation}

As explained in more detail in Sec.~\ref{subsec:human-guidance}, the X-axis of the $\boldsymbol{\Sigma}_{EE}$ and, consequently, the $\boldsymbol{\Sigma}_{H\ des}$ are always aligned with the direction of movement (see Fig. \ref{fig:software-architecture}).
Hence, the X-axis of the $\boldsymbol{\Sigma}_{H\ des}$ points to the direction of motion, and its Y-axis is parallel to the ground plane and orthogonal to the motion direction. 
Therefore, both the robot's torque gain ($\alpha_\mathrm{B}$) and the required adjustment of the end-effector's pose ($\mathrm{y_{A\ adaptive}}$) are calculated by using $\overrightarrow{{\mathrm{p}}^{sat}_x}$ and $\overrightarrow{{\mathrm{p}}^{sat}_y}$ which denote x and y components of the $\overrightarrow{\boldsymbol{p}^{sat}}$ vector, respectively.

For the calculation of $\alpha_\mathrm{B}$ the following function is used:
\begin{equation}\vspace{-2mm}
\alpha_\mathrm{raw} = - \frac{\overrightarrow{{\mathrm{p}}^{sat}_x}}{\mathrm{d}_{stop}} + 1
\end{equation}
\begin{equation} \label{eq:alpha_base_vel}\vspace{-2mm}
  \alpha_\mathrm{B} = 
  \begin{cases}\vspace{-2mm}
  1, \ \ \ \ \ \ \ \ \text{if} \ \ 1 \leq \alpha_\mathrm{raw} \\ \vspace{-2mm}
  \alpha_\mathrm{raw}, \ \ \ \ \text{if} \ \ 0 \leq \alpha_\mathrm{raw} \leq 1  \\ 0, \ \ \ \ \ \ \ \ \text{else}
  \end{cases},
\end{equation}
where $\mathrm{d}_{stop}$ is the saturation distance at which the robot base should not continue to move when the human lags behind.
The $\mathrm{y_{A\ adaptive}}$ is set equal to the $\overrightarrow{{\mathrm{p}}^{sat}_y}$. 

To achieve a lighter pull when the human deviation is low and a stronger pull as the deviation increases, a logistic function that adjusts lateral stiffness ($\mathrm{k_{\hat{y}}}$) is employed. This function modifies its output based on the $\mathrm{y_{A\ adaptive}}$ as follows:
\vspace{-2mm}
\begin{equation} \label{eq:ky}\vspace{-1mm}
    \mathrm{k_{\hat{y}}} = \zeta + \frac{(\eta - \zeta)}{1+e^{(\gamma-|\mathrm{y_{A\ adaptive}}|)\kappa}}
\end{equation}
where $\eta$, $\zeta$, $\gamma$, and $\kappa$  define the function's maximum and minimum asymptotes, the inflection point, and the steepness of the curve, respectively.

\subsection{Human guidance}\label{subsec:human-guidance}
\begin{algorithm}[!b]
	\caption{Human guidance algorithm}
	\label{alg:human-guidance}
	\begin{algorithmic}[1]
		\renewcommand{\algorithmicrequire}{\textbf{Input:}}
            \renewcommand{\algorithmicensure}{\textbf{Output:}}
		\REQUIRE $\boldsymbol{X}_{B\ targets}$, $\boldsymbol{X}_{B}$, $\mathrm{y_{A\ adaptive}}$
		\ENSURE  $\boldsymbol{X}_{B\ target}, \boldsymbol{\hat{X}}_{EE\ des}, \boldsymbol{X}_{H\ des}$
            \FOR{$target$ in $\boldsymbol{X}_{B\ targets}$}
                \IF {$EuclideanDistance(target, \boldsymbol{X}_{B}) \geq {D}_1$}
                    \STATE $\boldsymbol{X}_{B\ target} = target$
                    \STATE \textbf{break for}
                \ENDIF
            \ENDFOR
            \FOR{$odom$ in $\boldsymbol{X}_{B\ [0,...,t-1]}$}
                \IF {$EuclideanDistance(odom, \boldsymbol{X}_{B}) = {D}_2$}
                    \STATE $\boldsymbol{X}_{EE\ des} = odom$
                    \STATE $\boldsymbol{X}_{EE\ des}(z) =$ human hand height
                    \STATE \textbf{break for}
                \ENDIF
            \ENDFOR
            \STATE $\boldsymbol{\hat{X}}_{EE\ des} = \boldsymbol{X}_{EE\ des}$
            \STATE $\boldsymbol{\hat{X}}_{EE\ des}.translate(0,\mathrm{y_{A\ adaptive}},0)$
            \FOR{$EE$ in $\boldsymbol{X}_{EE\ des\ [0,...,t-1]}$}
                \IF {$EuclideanDistance(EE, \boldsymbol{X}_{EE\ des}) = {D}_3$}
                    \STATE $\boldsymbol{X}_{H\ des} = EE$
                    \STATE $\boldsymbol{X}_{H\ des}(z) =$ 2D laser height
                    \STATE \textbf{break for}
                \ENDIF
            \ENDFOR
		\RETURN $\boldsymbol{X}_{B\ target}, \boldsymbol{\hat{X}}_{EE\ des}, \boldsymbol{X}_{H\ des}$
	\end{algorithmic}
\end{algorithm}

The aim of this unit is to plan a trajectory that avoids any collision at the same time for (i) the mobile base, (ii) the robotic arm, and (iii) the human counterpart coupled with the robotic system. To do so, three desired poses need to be computed as depicted in Fig.\ref{fig:intro}, namely the base target pose $\boldsymbol{X}_{B\ target}$, the arm end-effector desired pose $\boldsymbol{\hat{X}}_{EE\ des}$, and the desired human pose at the legs height $\boldsymbol{X}_{H\ des}$. These poses are extracted from a set of poses where the previous component of the HRC framework has passed through as described hereafter. Alg.~\ref{alg:human-guidance} shows the pseudo-code of the illustrated method.

The next target waypoint for the mobile base $\boldsymbol{X}_{B\ target}$ is selected among the poses received in input by the \textit{obstacle avoidance} unit, i.e. $\boldsymbol{X}_{B\ targets}$, by finding the $target$ pose that is no less than a distance $D_1$ (see Fig.~\ref{fig:intro}) from the actual robot base pose $\boldsymbol{X}_{B}$ (lines 1-6). This pose is then given as input to the \textit{virtual torques shaping} unit to define the moving direction of the robot base.

Likewise, to ensure that the end-effector desired pose lies on the collision-free path, $\boldsymbol{X}_{EE\ des}$ is selected among the set of previous base poses $\boldsymbol{X}_{B\ [0,...,t-1]}$, by picking the value located at a distance $D_2$ (see Fig.~\ref{fig:intro}), that is set as the distance between the starting base and arm end-effector poses (lines 7-13). Since the pose for the robotic hand is selected among the base poses, and therefore does not lie at the same height, the z component of $\boldsymbol{X}_{EE\ des}$ is set to be equal to the human hand height. This pose is then translated on the y-axis (w.r.t. the arm base frame) by $\mathrm{y_{A\ adaptive}}$ (lines 14-15), to readjust the human pose if their are deviating from the imposed trajectory, as illustrated in Sec.~\ref{subsec:adaptive-pulling}.

In a similar manner, the human desired pose $\boldsymbol{X}_{H}$ is chosen as the previous desired end-effector poses vector $\boldsymbol{X}_{EE\ des\ [0,...,t-1]}$ that lies at distance $D_3$ (see Fig.~\ref{fig:intro}), i.e., the desired distance between the end-effector and the subject (lines 16-22). Since the human desired pose lies on a different horizontal plane, its height is adjusted to be on the same plane as the 2D laser scanners.

\subsection{Virtual torques shaping}\label{subsec:virtual-torques-shaping}
To define the motion direction and velocity for the mobile base, we can define the virtual torques to be given as input to the \textit{Cartesian admittance controller}. To do so, we define a vector $\boldsymbol{\overrightarrow{w}} = [x,y,\theta]^T$  that connects the robot base measured pose $\boldsymbol{X}_B$ and the next base target waypoint $\boldsymbol{X}_{B\ target}$ as:
\begin{equation}\vspace{-2mm}    
|\boldsymbol{\overrightarrow{w}}| = \sqrt{\boldsymbol{X}^B_{B\ target}(x)^2 + \boldsymbol{X}^B_{B\ target}(y)^2},
\end{equation}
\begin{equation}\vspace{-2mm}
\boldsymbol{\overrightarrow{w}}_x = \boldsymbol{X}^B_{B\ target}(x) / |\boldsymbol{\overrightarrow{w}}|,
\end{equation}
\begin{equation}\vspace{-2mm}
\boldsymbol{\overrightarrow{w}}_y = \boldsymbol{X}^B_{B\ target}(y) / |\boldsymbol{\overrightarrow{w}}|,
\end{equation}
\begin{equation}\vspace{-2mm}
\boldsymbol{\overrightarrow{w}}_\theta = atan2(-\boldsymbol{\overrightarrow{w}}_y, \boldsymbol{\overrightarrow{w}}_x).
\end{equation}

The virtual torques $\boldsymbol{\tau}^{vir}_B$ can be calculated by individually multiplying each component of the vector by a desired force $F_{des}$ and the gain $\alpha_{B}$. The desired force represents the maximum pulling force that the system can exert on the human operator, while the second parameter is computed as described in Sec.~\ref{subsec:adaptive-pulling}:
\begin{equation}\vspace{-2mm}
\boldsymbol{\tau}^{vir}_{B}(f_x) = \boldsymbol{\overrightarrow{w}}_x F^{des}_x \alpha_{B},
\end{equation}
\begin{equation}\vspace{-2mm}
\boldsymbol{\tau}^{vir}_{B}(f_y) = \boldsymbol{\overrightarrow{w}}_y F^{des}_y \alpha_{B},
\end{equation}
\begin{equation}\vspace{-1mm}
\boldsymbol{\tau}^{vir}_{B}(\mu_z) = \boldsymbol{\overrightarrow{w}}_\theta F^{des}_z \alpha_{B}.
\end{equation}

\section{Experiments and results}\label{sec:experiments}
\subsection{Experimental setup}\label{subsec:experimentalsetup}
\begin{figure}[!t]
\centering
\includegraphics[width=0.9\columnwidth]{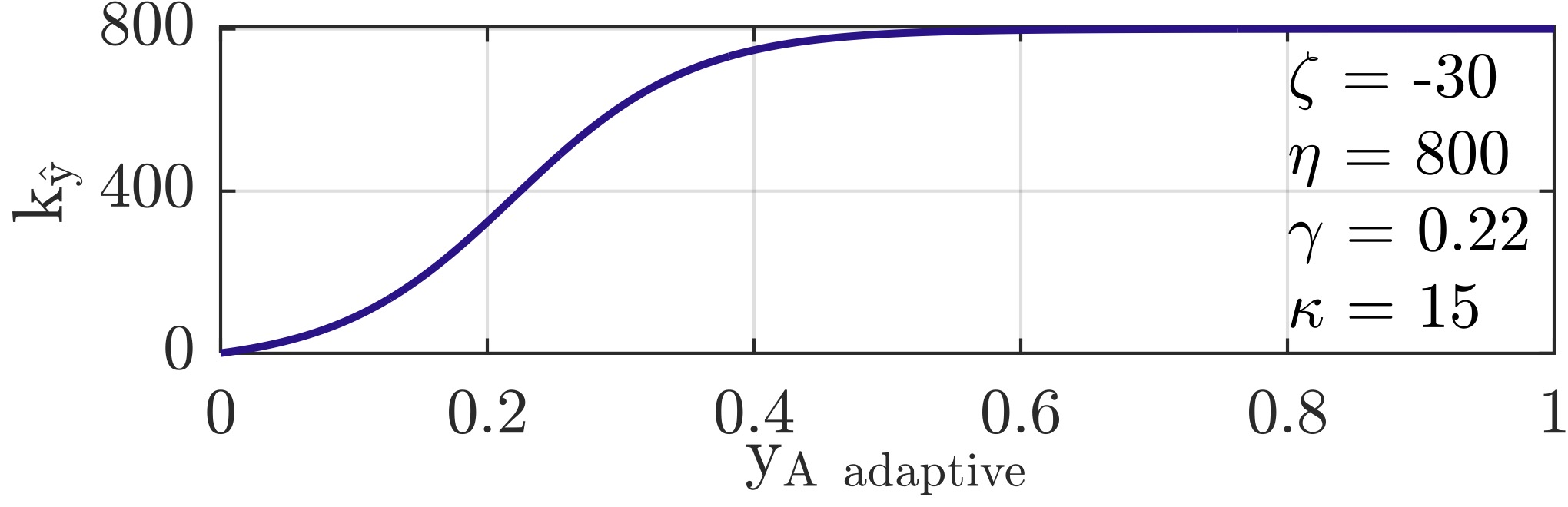}
\vspace{-3mm}
\caption{Lateral stiffness profile adopted by the \textit{adaptive pulling} unit.}
\label{fig:stiffness_profile}
\vspace{-7mm}
\end{figure}

We carried out two series of experiments deploying the mobile manipulator platform MOCA~\cite{wu2019teleoperation}. In the first one, we validated the framework, comparing results obtained with and without the \textit{adaptive pulling} method introduced in this study, conducted in a controlled laboratory environment (Sec.~\ref{subsec:experiments-validation}). In the baseline trials, i.e., without \textit{adaptive pulling}, the robot lateral stiffness $k_y$ was set to $0 N/m$ to leave freedom of motion to the subject. In the second experiment, we conducted a proof-of-concept experiment in a real-world scenario (Sec.~\ref{subsec:experiments-poc}). For both experiments, $k_x$ was set to $1000 N/m$ to maintain a rigid profile in the primary direction of motion, ensuring the human subject is pulled towards the desired path, and $k_z$ was set to $500 N/m$ to grant the user the option to make minor adjustments to the hand's vertical position. These stiffness values were selected within the range of typical human arm stiffness to ensure a natural and comfortable sensation during interactions \cite{ajoudani2018reduced}. $F^{des}$ was set to a value of $80 N$, that guarantees a reasonable final velocity for visually impaired subjects.
As can be seen from Fig.~\ref{fig:stiffness_profile}, the  \textit{adaptive pulling} unit generates lateral stiffness around 0 when the deviations are small, as in the baseline, whereas it escalates swiftly when the deviations increase during the task. A video of the experiments is available at \url{https://youtu.be/B94n3QjdnJE}.

\subsection{Framework validation}\label{subsec:experiments-validation}

\begin{figure}[!b]
\vspace{-5mm}
\centering
\includegraphics[width=0.9\columnwidth]{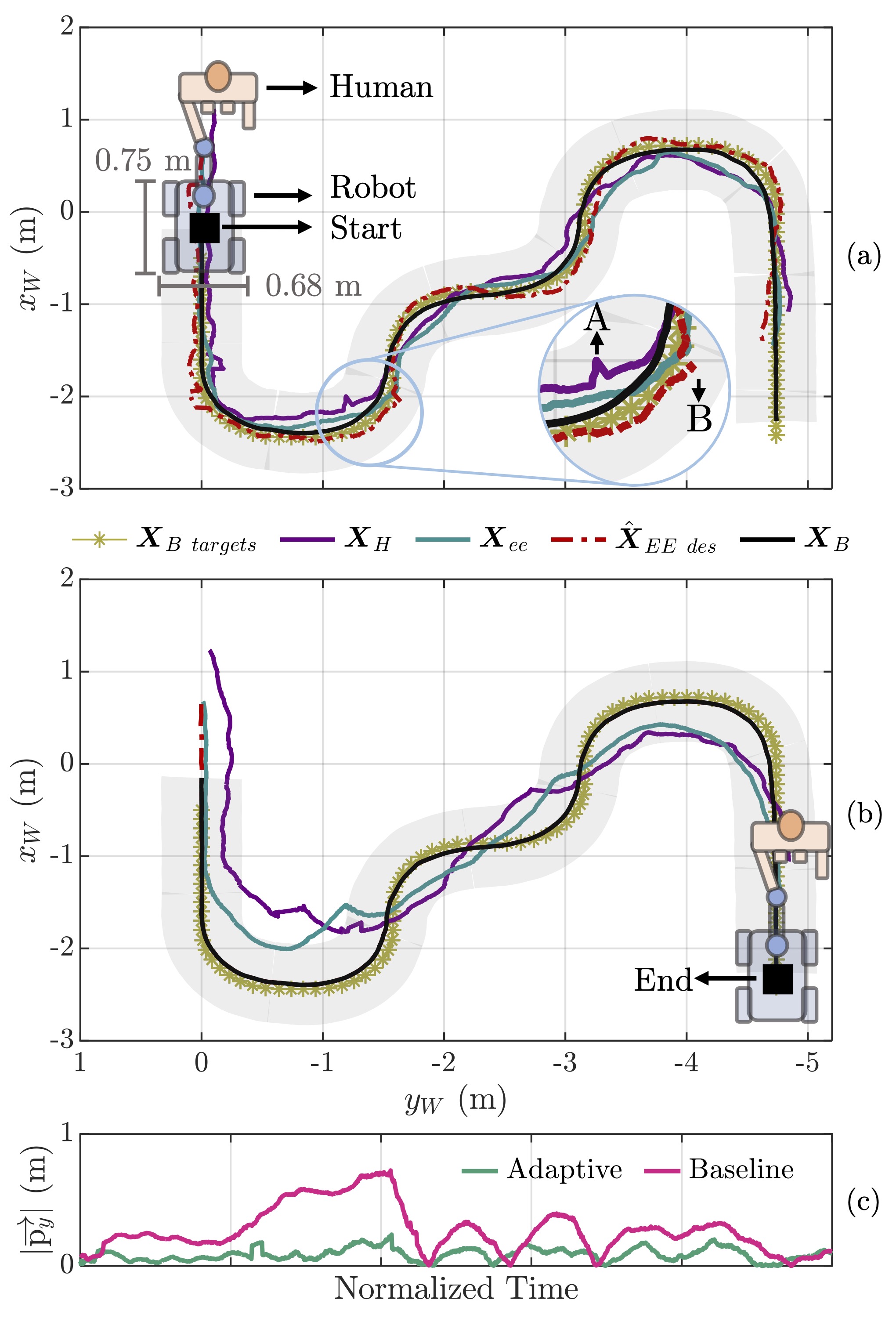}
\vspace{-4mm}
\caption{Plots on the XY plane using a predefined trajectory with adaptive pulling (a) and without (b). In the bottom, the variations of
 $|\overrightarrow{{\mathrm{p}}_y}|$ that reflect lateral deviations of the human from the desired path are presented (c).}
\label{fig:exp_validation_xy}
\end{figure}

To validate the presented framework, we first carried out an experiment, shaping a predefined trajectory, without employing the \textit{obstacle avoidance} planner. This approach was adopted to enable a more quantifiable comparison while using the \textit{adaptive pulling} unit and the baseline. This experiment was conducted by randomizing the trials and involved 12 healthy subjects (6 males, 6 females), whose sight was covered in order to simulate a visual impairment. Fig.~\ref{fig:exp_validation_xy} shows the trajectories in the XY plane followed by the HRC system with (top) and without (middle) using the \textit{adaptive pulling} planner. The $\boldsymbol{X}_{B\ targets}$ (dark yellow points) are tracked by the robot base pose $\boldsymbol{X}_{B}$ (black). The desired end-effector $\boldsymbol{X}_{EE\ des}$ and human $\boldsymbol{X}_{H\ des}$ poses are not plotted since they correspond to $\boldsymbol{X}_{B}$ shifted back in time as explained in Alg.~\ref{alg:human-guidance}, so they would not be visible. In the top plot it is possible to notice that, with the \textit{adaptive pulling}, $\boldsymbol{\hat{X}}_{EE\ des}$ (dashed red) tends to compensate the human deviations ($\boldsymbol{X}_{H}$ in purple) from the desired trajectory. For instance, in the focused portion of the plot we can observe that when the human is deviating on the robot's right (point A), thanks to the translation of $\boldsymbol{y}_{A\ adaptive}$ applied to $\boldsymbol{X}_{EE\ des}$, the robot end-effector desired pose $\boldsymbol{\hat{X}}_{EE\ des}$ stands on the robot left (point B) in  order to pull back the human on the desired collision-free path. By employing this method, the human pose $\boldsymbol{\Sigma}_H$ never extends beyond the robot's base footprint indicated by the gray area.  On the other hand, without using \textit{adaptive pulling} (middle plot), $\boldsymbol{\Sigma}_H$ goes beyond the robot's base footprint for 22.84\% of the time. This is not desirable since it can lead to potential collision with obstacles. The bottom plot highlights the lateral deviation of the human pose $\boldsymbol{\Sigma}_H$ w.r.t. their desired one $\boldsymbol{\Sigma}_{H\ des}$.

\begin{figure}[!t]
\centering
\includegraphics[width=0.93\columnwidth]{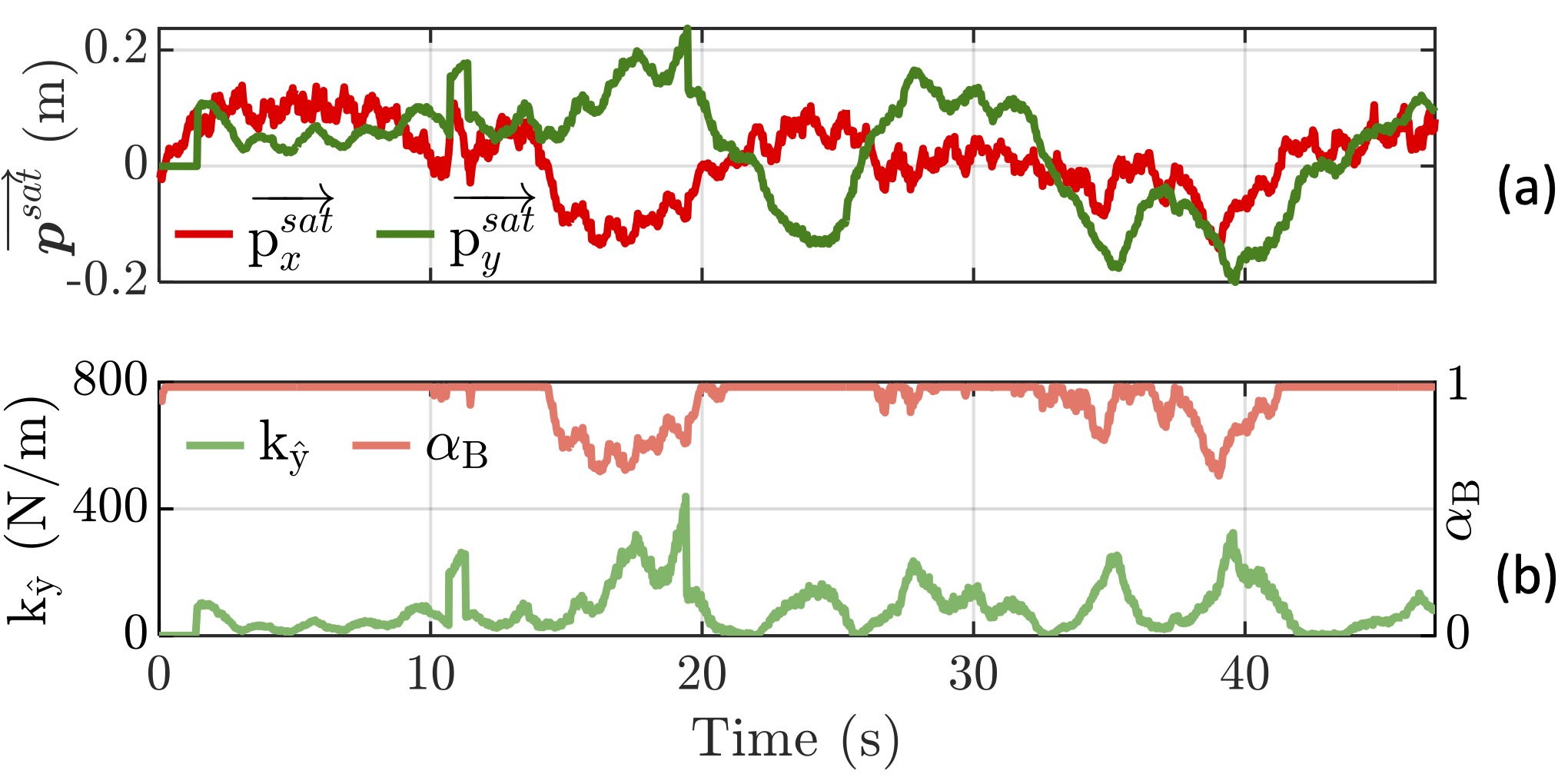}
\vspace{-4mm}
\caption{The $x$ and $y$ component of the $\overrightarrow{\boldsymbol{p}^{sat}}$ vector directly reflect the changes in the adaptive stiffness $\mathrm{k_{\hat{y}}}$ and base virtual torques gain $\alpha_{B}$.}
\label{fig:exp_validation_time}
\vspace{-6mm}
\end{figure}

\begin{figure}[!b]
\centering
\vspace{-4mm}
\includegraphics[width=0.93\columnwidth]{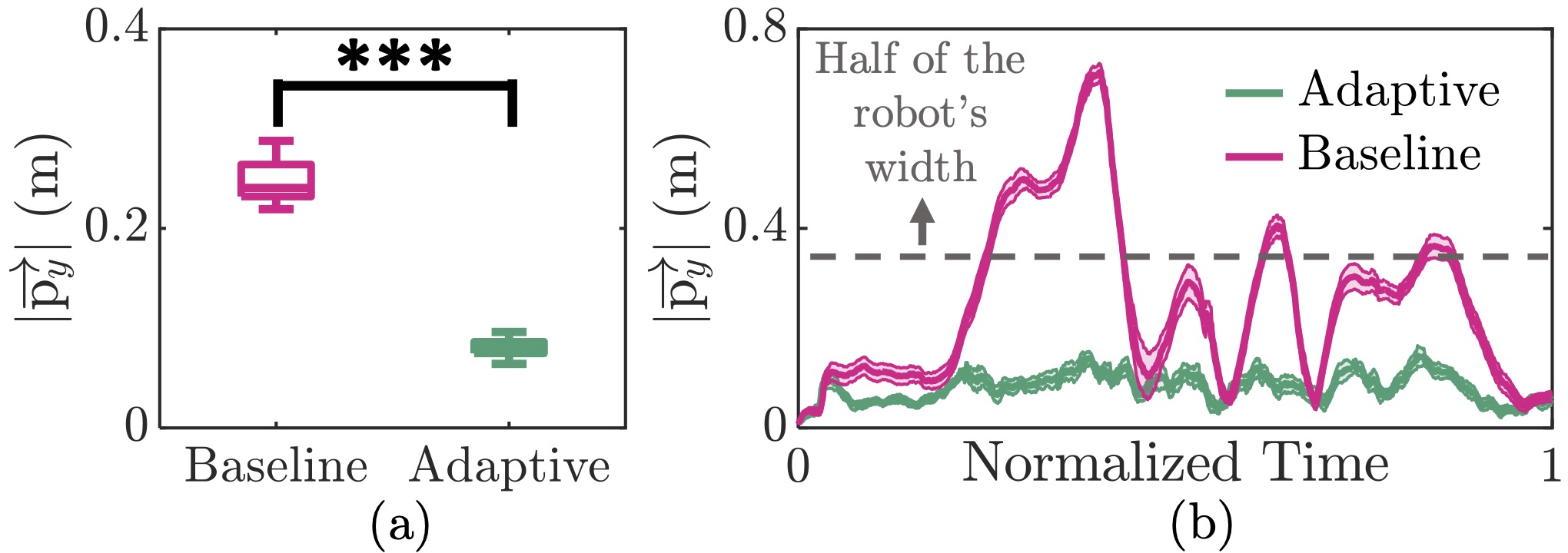}
\vspace{-4mm}
\caption{The average of $|\overrightarrow{{\mathrm{p}}_y}|$ values that reflect lateral deviations along with the outcome of the sign-test carried out: ***: p $<$ 0.001 (a) and the evolution of the $|\overrightarrow{{\mathrm{p}}_y}|$ throughout the trials (b) for both the cases with adaptive pulling and the baseline, considering all participants.}
\label{fig:deviation_combined}

\end{figure}

Fig.~\ref{fig:exp_validation_time} shows, for the \textit{adaptive pulling} trial illustrated above, the values of $\overrightarrow{\boldsymbol{p}^{sat}}$ (a), and the consequent adjustment of the lateral stiffness profile $\mathrm{k_{\hat{y}}}$ and base virtual torques gain $\alpha_{B}$ (b). It can be noticed that $\mathrm{k_{\hat{y}}}$ follows the trend of $\overrightarrow{{\mathrm{p}}^{sat}_y}$ according to~\eqref{eq:ky}, i.e., increasing the lateral stiffness profile when the subject deviates on the lateral direction, and that the gain $\alpha_{B}$ decreases below 1 whenever $\overrightarrow{{\mathrm{p}}^{sat}_x}$ goes below 0, i.e., giving rise to a reduction of the base velocity if the subject is lagging behind their desired pose. 
The small fluctuations in $\overrightarrow{{\mathrm{p}}^{sat}_x}$ stem from human footsteps since, as mentioned in Sec.\ref{subsec:legs-tracker}, the human pose is determined by averaging the legs’ centroids position.

Fig.~\ref{fig:deviation_combined} presents both the boxplots of average lateral deviations from the desired path ($|\overrightarrow{{\mathrm{p}}_y}|$) and the variations of the $|\overrightarrow{{\mathrm{p}}_y}|$ throughout the task as a function of normalized time for all the participants. 
As shown in Fig.~\ref{fig:deviation_combined}a, employing the adaptive pulling strategy led to a significant decrease in deviations compared to the baseline according to the sign-test (p $<$ 0.001). Moreover, when the deviation variations for the baseline trials are analyzed, it can be seen that the participants went out of the collision-free path, which can cause risky interactions with the environment (see Fig.~\ref{fig:deviation_combined}b).   
By using the proposed adaptive guidance framework, it was possible to keep the participants within safe boundaries.

\begin{figure}[!t]
\centering
\includegraphics[width=\columnwidth]{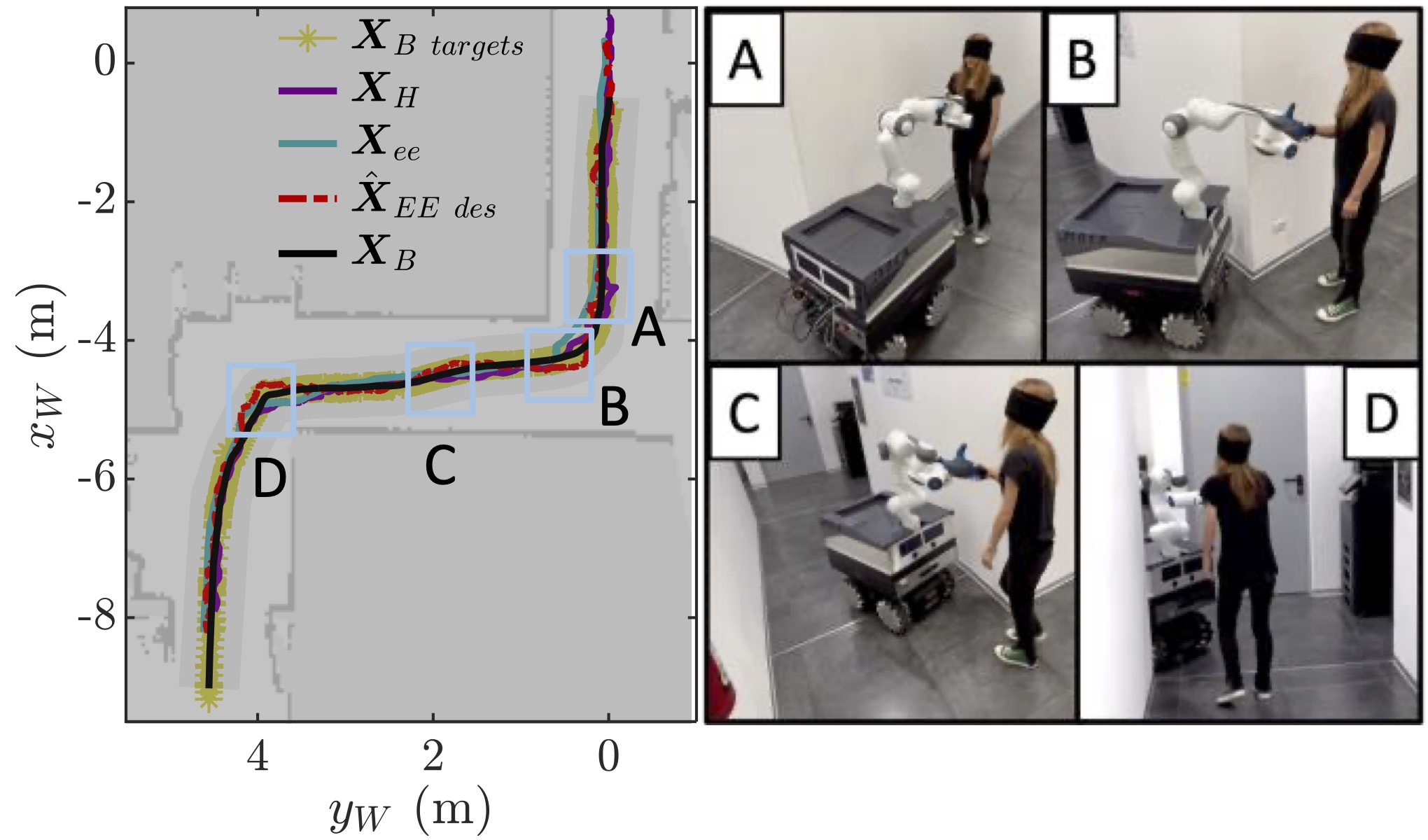}
\vspace{-6mm}
\caption{Real-world scenario: the \textit{obstacle avoidance} planner generates $\boldsymbol{X}_{B\ targets}$ poses and creates a map of the environment (left). The screenshots of four crucial moments [A-D] are shown on the right.}
\label{fig:move_base_exp}
\vspace{-6mm}
\end{figure}

\subsection{Real-world scenario proof-of-concept}\label{subsec:experiments-poc}
We also conducted a proof-of-concept experiment including the \textit{obstacle avoidance} planner to the framework. Fig.~\ref{fig:move_base_exp} shows on the left the trajectories in the XY plane similarly as in Fig.~\ref{fig:exp_validation_xy}. The plot is enriched with the map generated by the \textit{obstacle avoidance} unit. On the right side, we present four pivotal stages of the experiment conducted at points A, B, C, and D as depicted in the left plot. These stages demonstrate that even when encountering sharp 90-degree curves, all the components of the HRC framework successfully avoid the obstacle.

\vspace{-2mm}
\section{Conclusion and Discussion}
\label{sec:conclusion}
In this paper, we introduced a framework designed to guide individuals who are visually impaired through unfamiliar environments using a mobile manipulator. We achieve this by initially configuring a default compliant profile on the lateral axis to create a safer interaction between humans and the robot. However, when the individual strays from the intended trajectory, our system guides them back by adjusting impedance settings and directing the robotic arm. The experiments carried out with 12 subjects demonstrated the validity of the framework in avoiding obstacles. Future works will enhance the system robustness and user-friendliness by introducing a human-robot interface to communicate the desired destination, and by testing the framework with visually impaired subjects.

\newpage

\bibliographystyle{IEEEtran}
\bibliography{biblio}

\end{document}